\documentclass[10pt,twocolumn,letterpaper]{article}

\usepackage{cvpr}              

\usepackage[table,dvipsnames]{xcolor}
\usepackage{times}
\usepackage{epsfig}
\usepackage{graphicx}
\usepackage{amsmath}
\usepackage{amssymb}
\usepackage{tabularx}
\usepackage{booktabs}
\usepackage{dsfont}
\usepackage{multirow}
\usepackage{makecell}
\usepackage{arydshln}
\usepackage[usestackEOL]{stackengine}
\usepackage[accsupp]{axessibility}

\def\cvprPaperID{7688} 
\def\confName{CVPR}
\def\confYear{2022}

\include{math_notation}

\newif\ifdraft\draftfalse 

\ifdraft

\newcommand\hj[1]{\textcolor{DarkOrchid}{#1}} 
\newcommand\ds[1]{\textcolor{blue}{#1}} 
\newcommand\sm[1]{\textcolor{ForestGreen}{#1}} 

\else
\newcommand\hj[1]{#1}
\newcommand\ds[1]{#1}
\newcommand\sm[1]{#1}

\fi

\newcommand{\cellgray}{{\cellcolor[gray]{.95}}}

\usepackage[pagebackref=true,breaklinks=true,letterpaper=true,colorlinks,bookmarks=false]{hyperref}
\usepackage{xr-hyper}

\begin{document}


\title{Panoptic, {Instance} and Semantic Relations: A Relational Context Encoder to Enhance Panoptic Segmentation \vspace{-5pt}}


\author{
Shubhankar Borse \footnotemark[1]
\and
Hyojin Park \footnotemark[1]
\and
Hong Cai
\and
Debasmit Das
\and
Risheek Garrepalli
\and
Fatih Porikli\\
{Qualcomm AI Research \footnotemark[2] }\\
{\tt\small \{sborse, hyojinp, hongcai, debadas, rgarrepa, fporikli\}@qti.qualcomm.com}\\
}

	                              


\maketitle

\begin{abstract}

\vspace{-5pt}
This paper presents a novel framework to integrate both semantic and instance contexts for panoptic segmentation. In existing works, it is common to use a shared backbone to extract features for both things (countable classes such as vehicles) and stuff (uncountable classes such as roads). This, however, fails to capture the rich relations among them, which can be utilized to enhance visual understanding and segmentation performance. To address this shortcoming, we propose a novel Panoptic, Instance, and Semantic Relations (PISR) module to exploit such contexts. First, we generate panoptic encodings to summarize key features of the semantic classes and predicted instances. A Panoptic Relational Attention (PRA) module is then applied to the encodings and the global feature map from the backbone. It produces a feature map that captures 1) the relations across semantic classes and instances and 2) the relations between these panoptic categories and spatial features. PISR also automatically learns to focus on the more important instances, making it robust to the number of instances used in the relational attention module. Moreover, PISR is a general module that can be applied to any existing panoptic segmentation architecture. Through extensive evaluations on panoptic segmentation benchmarks like Cityscapes, COCO, and ADE20K, we show that PISR attains considerable improvements over existing approaches.

\end{abstract}

\vspace{-3pt}
\section{Introduction}
{
\let\thefootnote\relax\footnotetext{{
\hspace{-6.5mm} * These authors contributed equally. \\ \textdagger Qualcomm AI Research is an initiative of Qualcomm Technologies, Inc.}}}
\label{sec:introduction}
\vspace{-2pt}

Panoptic segmentation~\cite{kirillov2019panoptic} provides a unifying framework encompassing both semantic and instance segmentation. Its objective is to segment an image into \emph{things} and \emph{stuff}. Things include countable objects, such as cars and pedestrians, and stuff refers to uncountable concepts like sky and vegetation. Generating individual masks for \emph{things} is similar to instance segmentation, while predicting masks for \emph{stuff} is equivalent to performing semantic segmentation.  


\begin{figure}[t!]
\begin{center}
    \includegraphics[width=0.96\linewidth]{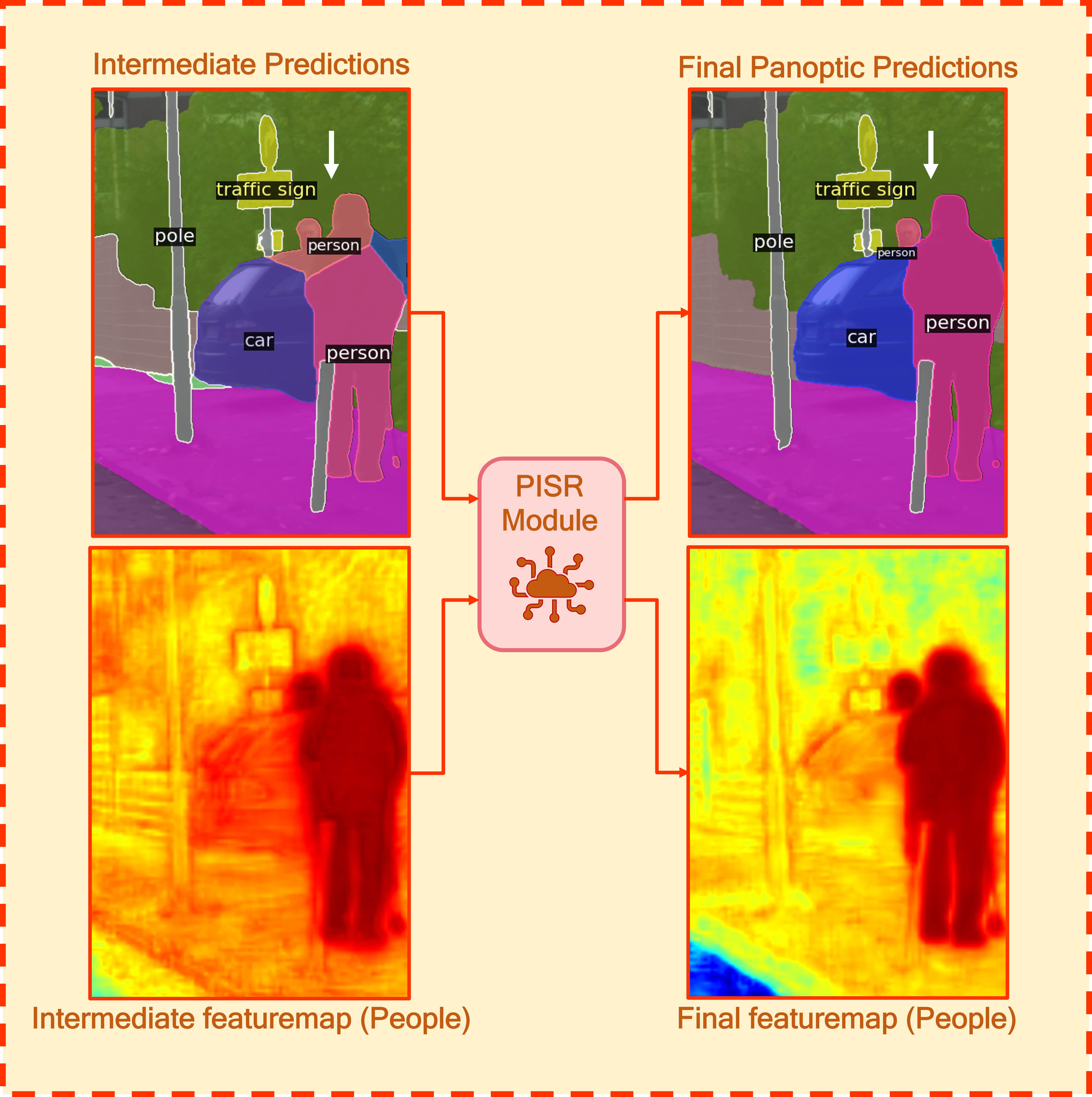}
\end{center}
\vspace{-15pt}
  \caption{Our proposed PISR module takes features and predictions for semantic and instance segmentation as inputs and applies a relational attention scheme. In this image, we show the input/output pairs obtained by applying PISR to a Panoptic-DeepLab model with ResNet50 backbone. The improvements in both the features and panoptic predictions are clearly visible. PISR produces more confident and accurate predictions in regions where a lot of instances are cluttered together (shown with white arrows). The feature maps are visualized for the persons class.}
\label{fig:teaser}
\vspace{-15pt}
\end{figure}



Since it aims to perform both tasks simultaneously, panoptic segmentation presents a challenge beyond conventional semantic or instance segmentation.
%
Early works~\cite{kirillov2019panoptic} propose to use separate modules for the two tasks, e.g., a Mask-RCNN module for instance segmentation and an FCN-based module for semantic segmentation. These two outputs are then combined during post-processing to generate panoptic segmentation. However, the accuracy in this case heavily relies on object detection quality. Besides, having two separate modules leads to redundant computations.  

Recently, \cite{li2021fully} proposed an architecture to process things and stuff together via a bottom-up, box-free approach. More specifically, semantic segmentation labels are first predicted, and then the instances are identified based on the grouped pixels. This architecture provides a unified approach, but it does not consider the relations among the semantic classes and instances. The object-contextual representations (OCR) module proposed by \cite{yuan2021segmentation} allows modeling the relations across semantic classes. However, it is designed for the semantic segmentation task and does not take instance information into account, which is critical for panoptic segmentation. For example, two images can have the same semantic classes, but their respective instances in the images can be very different. Our intuition is that in each semantic class, some instances can have drastically different visual appearances, sizes, and poses, while others may look similar. Therefore, understanding the relations (e.g., visual similarities) among the classes and instances will benefit panoptic segmentation. 

In this paper, we propose a novel \emph{Panoptic, Instance, and Semantic Relations (PISR)} module, which captures the key relations among semantic classes and instances for panoptic segmentation. Given an image, PISR computes encodings for semantic classes and pivotal instances. During this process, it automatically identifies on which instances it should focus more. PISR then applies attention to these encodings as well as global features to capture rich contextual relations that are useful for the final panoptic segmentation. PISR is a general component that can be used with any panoptic segmentation network, such as Panoptic-DeepLab~\cite{cheng2020panoptic}, Panoptic FPN~\cite{kirillov2019fpn}, and Maskformer~\cite{cheng2021maskformer}. To the best of our knowledge, this is the first time that relations among both semantic classes and object instances are exploited for panoptic segmentation \hj{explicitly}.

Our main contributions are summarized as follows:
\begin{itemize}
    \vspace{-7pt}
    \item We propose a novel Panoptic, Instance, and Semantic Relations (PISR) model to capture the relations among semantic classes and instances, providing richer contexts that enhance panoptic segmentation performance. 
    \vspace{-19pt}
    \item We devise a learnable scheme for PISR to automatically focus on more important instances while generating relational features. This provides a robust way to process the varying number of instances in each image.
    \vspace{-19pt}
    \item PISR is a universal module that can be used in any panoptic segmentation network.
    We show that it considerably enhances existing architectures on multiple datasets. We further conduct extensive ablation studies analyzing various aspects of PISR.
    \vspace{-10pt}
    \end{itemize}


\begin{figure*}[t]
\begin{center}
    \includegraphics[width=0.96\linewidth]{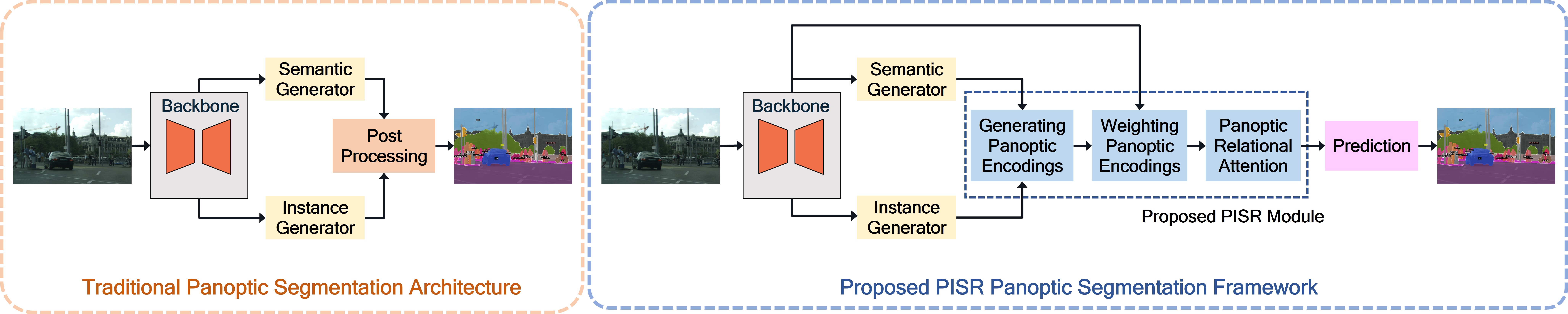}
\end{center}
    \vspace{-15pt}
   \caption{Left: Conventional panoptic segmentation architecture. Right: Our proposed Panoptic, Instance, and Semantic Relations (PISR) framework that can work with any base panoptic segmentation model.}
\label{fig:arch}
\vspace{-10pt}
\end{figure*}

\section{Related Work}
\vspace{-3pt}
\label{sec:related_work}
\textbf{Panoptic Segmentation:}
Aiming at assigning both instance and semantic labels to each image pixel, panoptic segmentation combines instance and semantic segmentation.\footnote{See~\cite{he2017mask, liu2018path, bolya2019yolact, gao2019ssap, sofiiuk2019adaptis, chen2019tensormask, lee2020centermask, wang2020solo, wang2020solov2, tian2020conditional, qi2021pointins} and \cite{long2015fully,chen2014semantic, zhang2022perceptual, zhao2017pyramid, ghiasi2016laplacian,chen2017deeplab,chen2019rethinking,wang2020deep,zhao2018psanet, huang2019ccnet,yuan2021segmentation, borse2021hs3, borse2021inverseform, zhang2022auxadapt} for recent methods on instance and semantic segmentation respectively, \cite{Cai2021xdistill} for using segmentation for other tasks, and \cite{minaee2021image} for an extensive survey.}
One of the earliest methods for panoptic segmentation~\cite{kirillov2019panoptic} utilizes predictions from separate instance and semantic segmentation models, but using it is inefficient as the two models do not share parameters. Recent approaches can be grouped into two categories: 1) separate representations and 2) unified representation. In the first case, instances and semantic classes are segmented using a single model, but through two different branches. Things are segmented either via boxes~\cite{yang2020sognet,li2020unifying,chen2020banet} or by using box-free methods~\cite{yang2019deeperlab}. Stuff is usually segmented using a fully convolutional branch. Other methods that use separate representations include AUNet~\cite{li2019attention}, Panoptic FPN~\cite{kirillov2019fpn}, and UPSNet~\cite{xiong2019upsnet}. In contrast, unified representation approaches segment things and stuff based on features generated from shared layers~\cite{li2018weakly,li2021fully}. However, these methods do not consider the contextual/relational information of things and stuff when generating features and making predictions.

\textbf{Relational Contexts in Segmentation: }
More recently, researchers have started to look into learning and exploiting relational information for segmentation. \cite{zhao2018psanet, huang2019ccnet,yuan2021segmentation,cheng2021maskformer} utilize an attention mechanism to capture pixel inter-dependencies over an image. Specifically, in~\cite{yuan2021segmentation}, the authors propose Object-Contextual Representations (OCR), which captures the correlations between pixel-wise features and object region representations and has enabled state-of-the-art performance in semantic segmentation. However, OCR only considers entire class regions and cannot be used for object instances. In our work, we go beyond OCR and propose a novel method, PISR, to learn and capture relations among both object instances (things) and stuff. Our proposed method can be used with any existing panoptic segmentation architecture and can considerably enhance the segmentation performance, as we shall see in the paper.
\hj{
In\cite{cheng2021maskformer}, the authors build implicit relations based on transformer models using queries.
Different from this, PISR is a general module to capture explicit relations among panoptic classes and the relations between each class and the scene. 
Furthermore, it reweights each component to focus on the more useful information and enhance any existing model, including transformer-based ones.}

\vspace{-3pt}
\section{Proposed Approach}
\label{sec:method}
\vspace{-3pt}
This section describes our novel Panoptic, Instance, and Semantic Relations (PISR) model that captures the relations from semantic and instance features to generate a final representation that is more informative for panoptic segmentation. We provide an overview of PISR and how it works with any given base architecture (e.g., an existing panoptic segmentation model) in Section~\ref{subsec:arch}, and then discuss its components in detail,  including how PISR generates the initial panoptic encodings, reweights them, and applies attention to capture key correlations, in Sections~\ref{subsec:th_st_gen},~\ref{subsec:weighting}, and~\ref{subsec:PISR}.


\subsection{Applying PISR for Panoptic Segmentation}
\label{subsec:arch}
\vspace{-2pt}
A general panoptic segmentation architecture usually consists of four parts: 1) a backbone for feature extraction, 2) a semantic generator that outputs semantic segmentation, 3) an instance generator that outputs instance segmentation, and 4) a post-processing block that combines the two types of segmentations to produce the final panoptic segmentation. An illustration of this is given in Figure~\ref{fig:arch}~(left). We refer to such a panoptic segmentation pipeline as a base architecture, which encompasses most existing methods.

PISR is a universal module and can be appended to any panoptic segmentation architecture. This not only includes traditional architectures \cite{cheng2020panoptic, kirillov2019fpn, li2021fully} but also recent state-of-the-art models \cite{cheng2021maskformer, wang2021max}.
First, it takes the outputs from the semantic (stuff) and instance (thing) generators. Only the top $K$ most confident predictions are used to confine the instances to more reliable ones. We refer to each semantic class and each selected instance in these outputs as a panoptic category (e.g., the car class, person~1, person~2). PISR generates an initial encoding for each panoptic category, which summarizes the key features of the pixels assigned to that category. It then automatically reweights these initial encodings to highlight the more important ones. Next, the weighted panoptic encodings are fed into a Panoptic Relational Attention module, after which the enhanced features are sent to the prediction stage to generate the final panoptic segmentation. These steps are illustrated in Figure~\ref{fig:arch}~(right). 


When we do end-to-end training of an architecture with PISR, we apply the usual semantic and instance segmentation losses to the final estimated outputs. In addition, we also exert semantic and instance losses to the intermediate outputs from the semantic and instance generators that come with the base architecture. More formally, our total training loss function can be written as follows: \vspace{-5pt}
\begin{equation}
    \label{eq:loss}
     \mathcal{L} = \mathcal {\gamma}\cdot ( \mathcal{L}'_{sem} + \mathcal{L}'_{ins} ) + \mathcal{L}_{sem} + \mathcal{L}_{ins}, \vspace{-5pt}
\end{equation}
where $\mathcal{L}_{sem}$ and $\mathcal{L}_{ins}$ are the loss functions for predicting final instances and semantic segmentation respectively, $\mathcal{L}'_{sem}$ and $\mathcal{L}'_{ins}$ are the intermediate semantic and instance loss functions and $\gamma$ is treated as a hyperparameter. For both intermediate and final supervisions, we apply the same loss functions used to train each base architecture as reported in their original paper. For instance, when using Panoptic-DeepLab~\cite{cheng2020panoptic} as the base model, the semantic segmentation is supervised by a cross-entropy loss while the instance masks are supervised by an MSE loss for center heatmaps and an L1 loss for offsets. Training details and hyperparameters for each experiment are discussed in the Appendix.

\begin{figure*}[t!]
\begin{center}
    \includegraphics[width=1\linewidth]{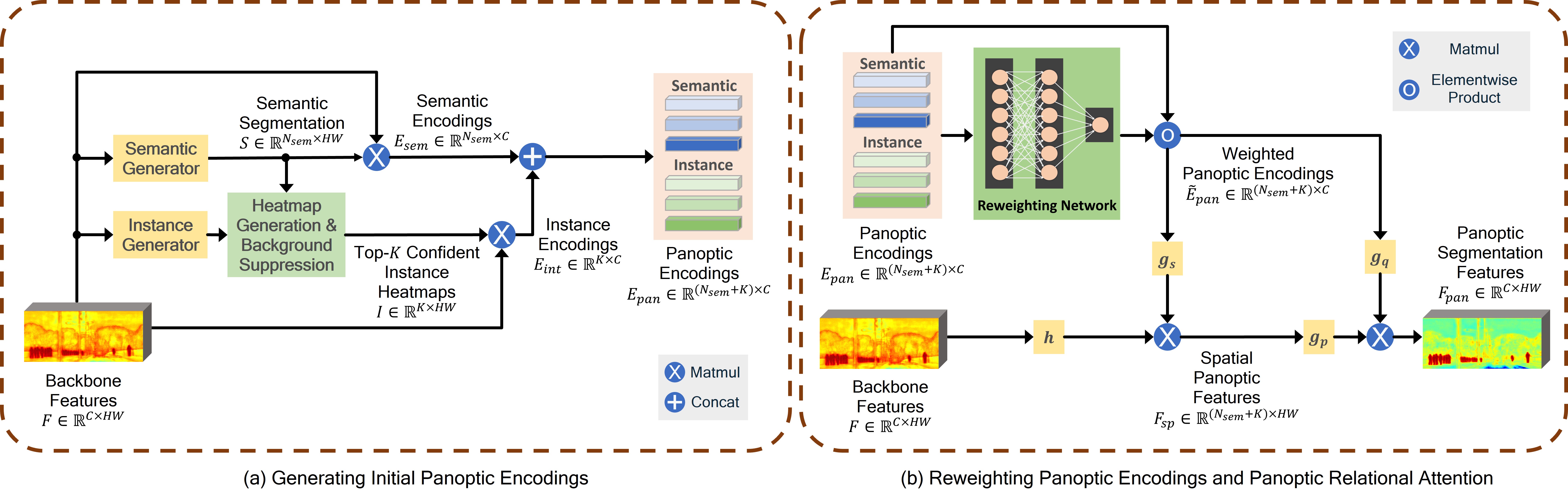}
\end{center}
\vspace{-17pt}
  \caption{(a) Details on how the initial panoptic encodings are generated. (b) Details on how PISR reweights the initial encodings and subsequently applies two-stage attention by employing PRA to generate the final panoptic segmentation features.}
\label{fig:pisr_details}
\vspace{-10pt}
\end{figure*}

\subsection{Generating Initial Panoptic Encodings}
\label{subsec:th_st_gen}
\vspace{-2pt}
Given the features supplied by the backbone, we generate panoptic encodings which summarize the key features of the semantic classes and selected instances. The procedures for generating panoptic encodings are shown in Figure~\ref{fig:pisr_details}~(a) and described in detail in the following.

\textbf{Semantic Encodings: }
Suppose that the backbone network generates a feature map $F \in \mathbb{R}^{C\times HW}$, where $C$, $H$, and $W$ are the number of channels, height, and width of the feature map. Taking $F$ as input, the semantic generator produces a soft semantic segmentation map $S \in \mathbb{R}^{N_{sem} \times HW}$, where $N_{sem}$ is the number of semantic classes, and for each pixel location, a probability vector that indicates how likely this pixel belongs to different classes. We calculate the semantic encodings $E_{sem} \in \mathbb{R}^{N_{sem}\times C}$ by multiplying $S$ and $F$: $E_{sem} = SF^T$. These encodings contain the most prominent features for the semantic classes.

\textbf{Instance Encodings:}
Standard instance predictions contain a center mass $M \in \mathbb{R}^{1\times HW}$ and a center offset $O \in \mathbb{R}^{2\times HW}$. $M$ is the objectness score, which we use to select the top-$K$ most confident center locations as shown in Figure~\ref{fig:ocr_pisr} (e), (f).
Given these $K$ selected centers, we then produce $K$ \hj{initial} heatmaps based on their center offsets.\footnote{\hj{We generate the initial instance heatmap, $H_{init}\in \mathbb{R}^{K\times HW}$, by performing simple instance center regression:
$H_{inst}(i,j)=1 - C(M(i,j) - (i+O_x(i,j),j+O_y(i,j)))$, where $C$ is a normalization constant and $(i,j)$ is the pixel location. We provide further visualizations for these initial instance heatmaps in the Appendix.
}}
Next, we convert the predicted semantic segmentation $S$ into a binary segmentation of things and stuff, and then multiply it with the \hj{initial} heatmaps in order to suppress the background. The resulting instance heatmaps are denoted as $I\in \mathbb{R}^{K\times HW}$.
Finally, we calculate the instance encodings $E_{ins}\in \mathbb{R}^{K\times C}$ by multiplying $I$ and $F$: $E_{ins} = IF^T$.

\textbf{Panoptic Encodings:} 
The semantic encodings and instance encodings are concatenated to form the final panoptic encodings: $E_{pan} \in \mathbb{R}^{(N_{sem} + K) \times C}$. Each panoptic encoding then summarizes the key features of a semantic class or one of the selected instances.

\vspace{-4pt}
\subsection{Reweighting Panoptic Encodings}\label{subsec:weighting}
\vspace{-2pt}
Given the panoptic encodings $E_{pan}$, we further reweigh them based on their importance. More specifically, we use a small 2-layer fully-connected network with a sigmoid output layer to generate the weights.\footnote{Larger reweighting networks provided no significant improvements.} It takes $E_{pan}$ as input and outputs the weight vector $\omega \in \mathbb{R}^{(N_{sem}+K)\times 1}$. Each element in $\omega$ is a predicted importance score for a panoptic category. 
\ds{We then compute the weighted panoptic encodings as follows:  $\tilde{E}_{pan} = E_{pan} \circ [\omega]$, where $[\omega] \in \mathbb{R}^{(N_{sem}+K)\times C}$ is the broadcasted version of $\omega \in \mathbb{R}^{(N_{sem}+K)\times 1}$ across the $C$ dimension of $E_{pan}$ while $\circ$ is the element-wise product.}


By doing this, PISR learns to focus on the important semantic classes and instances, while suppressing the less relevant ones. Although this reweighting network employs a simple structure, it provides a critical capability for PISR to be more robust to $K$. As we shall see in the experiments, as $K$ increases, this reweighting allows PISR to generate improving panoptic segmentation whereas the performance can degrade without weighting. \sm{By analyzing output weights, we found that the reweighting network learns to down-weight classes that are not present in the scene and false-positive instances that make it into top-$K$ predictions.}

\subsection{Panoptic Relational Attention}
\label{subsec:PISR}
\vspace{-2pt}

\begin{figure*}[t]
\center
\includegraphics[ width=0.96\linewidth]{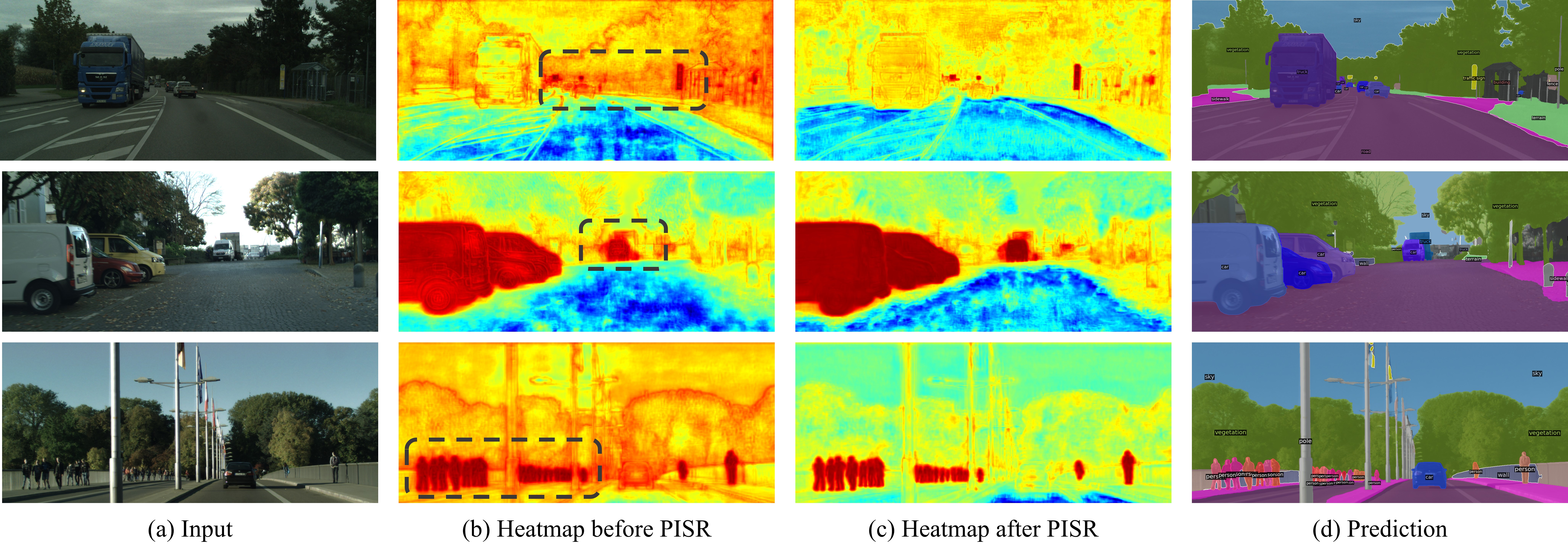}
\vspace{-10pt}
\caption{Visualization of the panoptic segmentation features before and after PISR, for three sample classes (top: traffic sign, middle: car, bottom: person). In the four columns we show: (a) input RGB image, (b) heatmap visualization for a given class before PISR, (c) heatmap visualization for the same given class after PISR, and (d) final panoptic segmentation. Darker red (lighter blue) colors indicate stronger (weaker) signals in the feature map. With PISR, shapes of the target-class objects are more accurately captured and their signals are considerably stronger. In regions where multiple instances interact with each other, PISR also captures clearer features  as indicated by the sharp boundaries of instances such as people and cars. Sample improvements are highlighted by the boxes.
}
\label{fig:Act_map}
\vspace{-10pt}
\end{figure*}

Panoptic segmentation requires a holistic understanding of the scene, including both things and stuff. However, existing approaches do not fully utilize the relations among semantic classes and instances. To enable the network to leverage the underlying relational contexts, we devise a Panoptic Relational Attention (PRA) module computing correlations across the panoptic categories based on panoptic encodings. PRA takes global features $F$ and panoptic encodings $\tilde{E}_{pan}$ as input. Two stages of attention are then applied to extract various types of correlations. The details of PRA are shown in Figure~\ref{fig:pisr_details}~(b) and discussed as follows.

First, we correlate the weighted panoptic encodings with the spatial features. This produces a spatial panoptic feature map: $F_{sp} = g_{s}(\tilde{E}_{pan})^T h(F)$, where $F_{sp}\in \mathbb{R}^{(N_{sem} + K)\times HW}$, and $g_{s}$ and $h$ contain $1\!\!\times\!\!1$ and $3\!\!\times\!\!3$ convolutional layers, respectively. This captures the panoptic signals in each pixel location. Next, we take the spatial panoptic feature map $F_{sp}$ and correlate it with the weighted panoptic encodings $\tilde{E}_{pan}$. This produces the final panoptic segmentation features: $F_{pan} = g_q(\tilde{E}_{pan})^T g_p(F_{sp})$, where $F_{pan}\in \mathbb{R}^{C\times HW}$ and $g_p$ and $g_q$ contain $1\!\!\times\!\!1$ convolutional layers. This final feature map $F_{pan}$ carries the enhanced panoptic signals over the spatial pixel locations and is fed to the final prediction stage to generate the semantic and instance segmentation, and the final panoptic segmentation.

\vspace{-3pt}
\section{Experiments}
\label{sec:experiments}
\vspace{-2pt}
In this section (and also in the Supplementary File), we present comprehensive performance evaluations of PISR on large benchmark datasets, compare it with baselines and the current state of the art (SOTA), and conduct extensive ablation studies on various aspects of PISR.


\vspace{-2pt}
\subsection{Experimental Setup}
\vspace{-3pt}
\textbf{Datasets:} We evaluate performance on Cityscapes~\cite{cordts2016cityscapes}, COCO~\cite{lin2014microsoft}, and ADE20K~\cite{zhou2017scene} panoptic segmentation datasets. Cityscapes consists of 3,475 annotated images of size $1024\! \times \! 2048$, with 2,975 training and 500 validation images. It covers 11 stuff and 8 thing classes.  COCO consists of 118,000 training, 5,000 validation, and 20,000 test images, covering 53 stuff and 80 thing classes. ADE20K contains 20,210 training, 2,000 validation and 3,000 test images, with 35 stuff and 115 thing classes.

\textbf{Networks and Training:} We implement PISR with several state-of-the-art base panoptic segmentation architectures including Panoptic-DeepLab~\cite{cheng2020panoptic}, Panoptic-FPN~\cite{kirillov2019fpn}, and Maskformer~\cite{cheng2021maskformer}. We experiment with various backbones for these models, including ResNet-50, ResNet-101~\cite{he2016deep}, HRNet-w48~\cite{wang2020deep}, Swin-L~\cite{liu2021swin} and ResNet-50-FPN~\cite{kirillov2019fpn}. When training each base architecture with PISR, we use the original semantic and instance loss functions for both intermediate and final supervisions. We follow the original training settings, e.g., hyperparameters, training schedule, etc. Training details and hyperparameters are included in the supplementary material.

\textbf{Baseline:} In addition to comparing with the existing panoptic segmentation models, we also implement a strong baseline where the Object-Contextual Relationship (OCR) module~\cite{yuan2021segmentation} is applied to Panoptic-DeepLab. By comparing with this baseline, we directly show the advantage of leveraging both things and stuff relations in PISR, as compared to OCR which only considers semantic classes.

\textbf{Evaluation Metrics:} We use standard metrics to evaluate panoptic segmentation performance, including panoptic quality (PQ)~\cite{kirillov2019panoptic}, semantic quality (SQ), and recognition quality (RQ). We further report PQ scores for things and stuff, denoted as PQ$^{th}$ and PQ$^{st}$, respectively. For semantic segmentation and instance segmentation predictions, we report mean Intersection over Union (mIoU) and Mask Average Precision (AP)~\cite{lin2014microsoft} scores, respectively.

\vspace{-3pt}
\subsection{Heatmap Visualization} 
\vspace{-3pt}
We show the visualization of the enhanced heatmaps by using our proposed PISR (column (c)), as well as the original ones without PISR (column (b)) in Figure~\ref{fig:Act_map}. For each row, we visualize the heatmaps for a sample class. The heatmaps of all the selected instances of the same class are overlaid on the same image for better visualization. 

It can be seen that PISR significantly improves the recognition of the target classes, as shown by the stronger signals depicted by darker red colors. We also observe that PISR recognizes sharper boundaries around regions where lots of instances are interacting with each other.

\begin{figure}[t]
\begin{center}
    \includegraphics[width=1\linewidth]{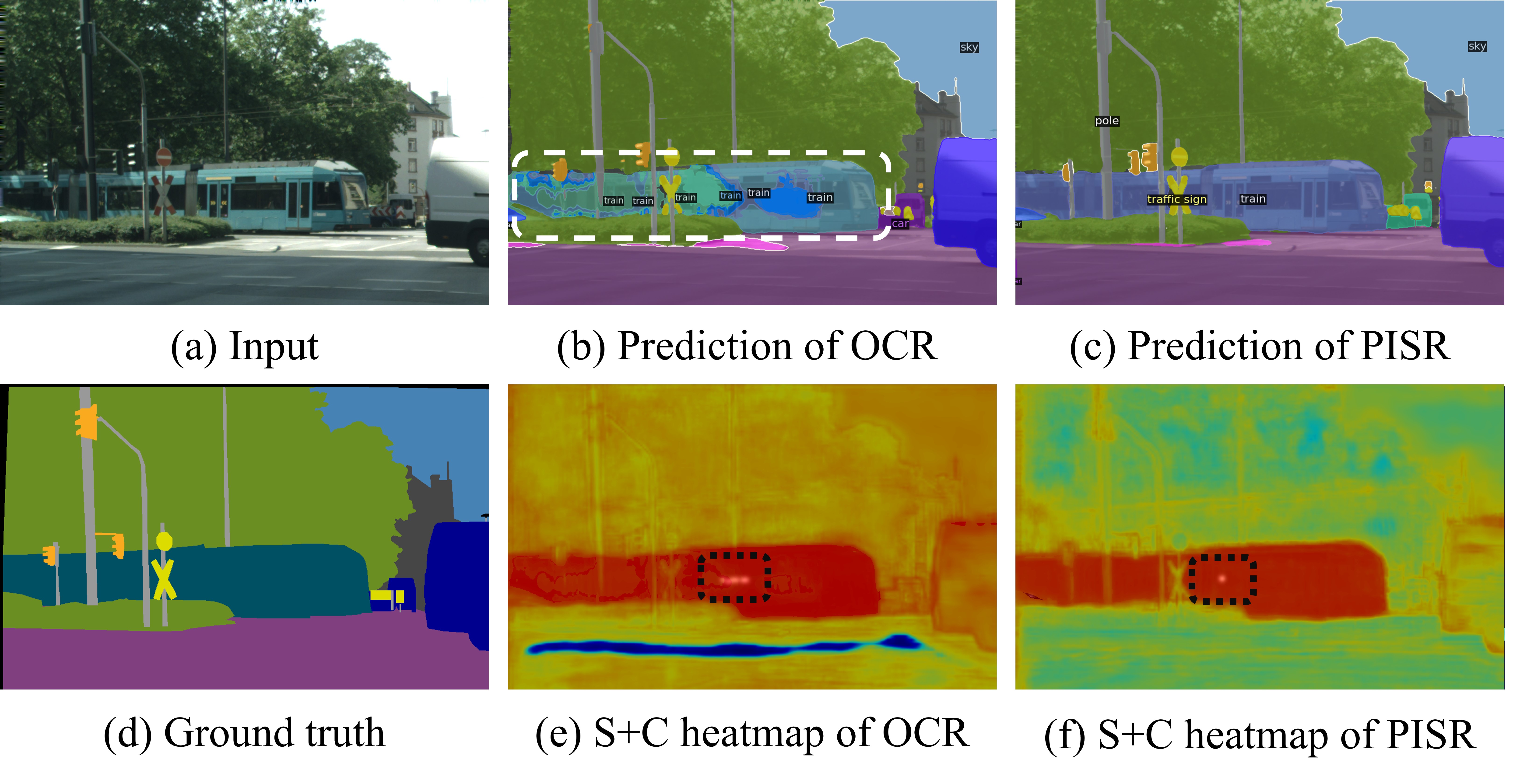}\\
\end{center}
    \vspace{-20pt}
  \caption{ Visualization of the comparison between OCR and PISR. 
  For (e) and (f), we visualize semantic heatmaps and instance center results overlaid together for the ``train" class.
  Note that OCR predicts multiple overlapped centers, even when the number of instances of ``train" class is only one.  OCR does not consider instance features, hence fails to correct wrong prediction of instances.
  PISR on the other hand encompasses every semantic and instance feature by analyzing similarity among them, and thus successfully produces better instance masks and centers. 
 }
\label{fig:ocr_pisr}
\vspace{-10pt}
\end{figure}

In Figure~\ref{fig:ocr_pisr}, we show visualizations for two Panoptic-DeepLab-ResNet50 models, trained with OCR and PISR, respectively. We overlay the obtained Semantic and Center (S+C) heatmaps for the ``trains" category and present them together. We also show the final predictions obtained from both models.
It is visible that OCR and PISR both generate similar semantic heatmaps; however, OCR predicts multiple centers for the ``train'' category compared to PISR, leading to poor instance segmentation results. In this case, even though both intermediate predictions are similar, PISR is able to effectively rectify the errors by incorporating instance information into its relational attention scheme.

\subsection{Evaluation Results}
\begin{figure}[t]
\center
\includegraphics[width=1\linewidth]{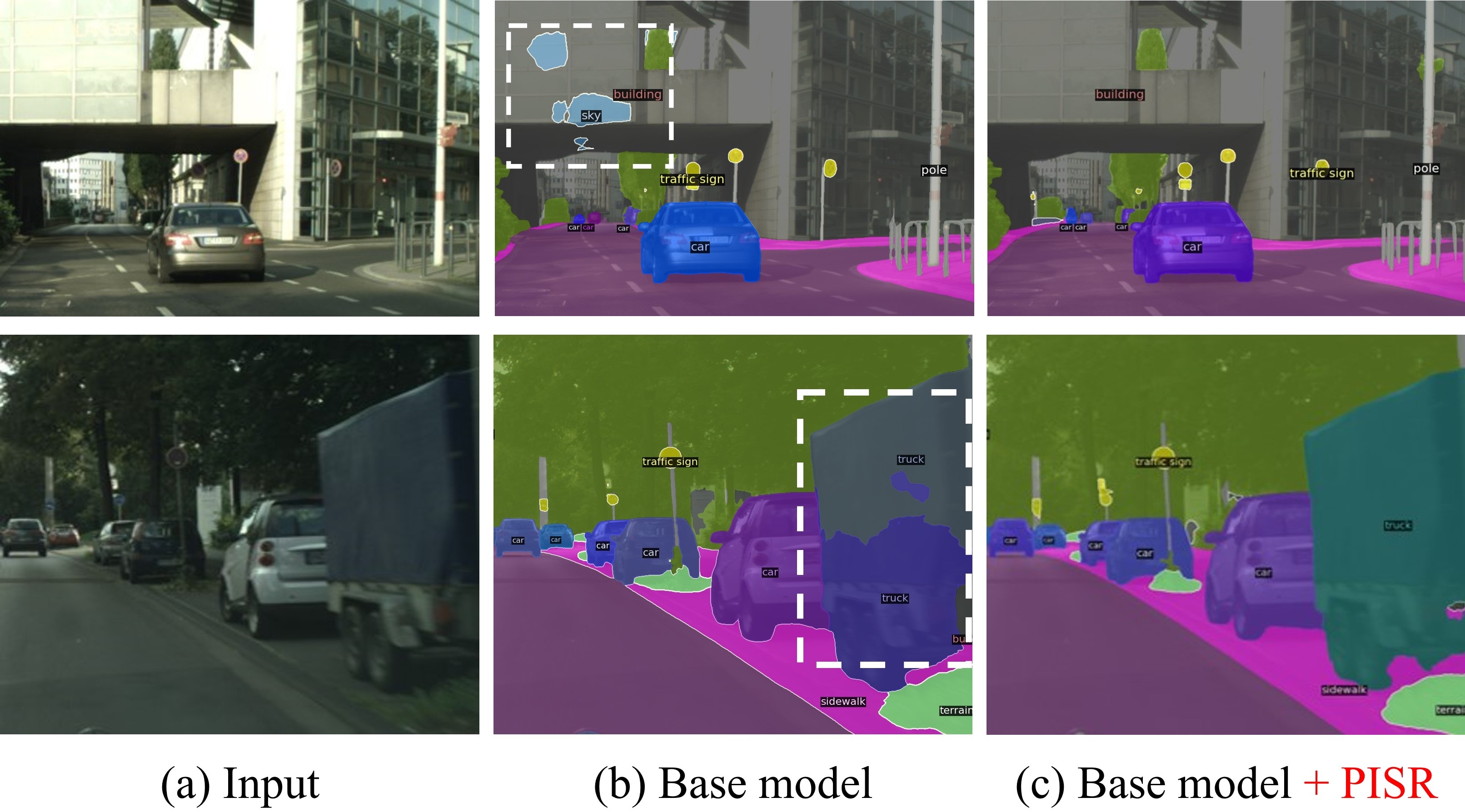}
\vspace{-17pt}
\caption{
Qualitative results \textbf{on Cityscapes:}
(a) Input images.
(b) Predictions by using Panoptic-DeepLab (ResNet-50).
(c) Our results by applying PISR to Panoptic-DeepLab (ResNet-50).
The overall panoptic segmentation quality improves with PISR. Dashed boxes highlight sample regions where PISR significantly enhances the baseline prediction.}
\label{fig:Visualcity}
\vspace{-13pt}
\end{figure}

\vspace{-2pt}
\textbf{Cityscapes:} We report results on the Cityscapes-val split in Table~\ref{tab:city} and Figure~\ref{fig:Visualcity}. We use Panoptic-DeepLab with various backbones as our base model, and train these networks using PISR. In the first part of Table~\ref{tab:city}, we compare with recent methods which use ResNet-50 or its variants as the backbone. It can be seen that our proposed PISR module significantly improves panoptic segmentation performance of the base model and also outperforms other pipelines. PISR enables much higher gains over the original Panoptic-DeepLab as compared to the baseline where OCR is used. This is due to the fact that PISR considers the relations among both semantic classes and key instances, while OCR only utilizes semantic relations.

In the second section of Table~\ref{tab:city}, we compare with SOTA methods that use larger backbones (e.g., ResNet-101, EfficientNet). In this case, we use the HRNet-w48 backbone. Our PISR-enhanced Panoptic-DeepLab outperforms the SOTA methods. For fair comparison, we consider versions of these networks initialized using ImageNet pre-trained weights and we perform single-scale inference.\footnote{Multi-scale inference is a computationally expensive technique that can in principle be applied to any network, including a PISR-enhanced model. In our evaluation, we do not apply multi-scale inference to ensure a fair comparison with all recent works which do not use it.}

Figure~\ref{fig:Visualcity} shows qualitative results obtained by using Panoptic-DeepLab (ResNet-50) without and with PISR. It can be clearly seen from the highlighted regions that using PISR leads to overall more accurate segmentation as compared to the baseline prediction. For instance, PISR successfully exploits panoptic relations to predict the correct semantic label for ``building" (and not ``sky") even though it contains a reflection of the sky. 

\begin{table}[t]
\scriptsize
\centering
\begin{tabular}{lc|ccc}

\Xhline{2\arrayrulewidth}
\textbf{Method}        & \textbf{Backbone} & \textbf{PQ}   & \textbf{AP}   & \textbf{mIoU} \\
\hline
Panoptc FPN \cite{kirillov2019fpn}      & RN50-FPN  & 58.1 & 33.0   & 75.7 \\
UPSNet  \cite{xiong2019upsnet}           & RN50-FPN  & 59.3 & \textbf{33.3}   & 75.2 \\
Unifying \cite{li2020unifying}           & RN50-FPN  & 61.4  & \textbf{33.3}  & 79.5    \\

LPSNet \cite{Hong_2021_CVPR}              & RN50-FPN  & 60.4  & 33.0  & 78.6    \\
Seamseg \cite{Porzi_2019_CVPR}            & RN50-FPN  & 60.2  & \textbf{33.3}  & 74.9  \\
COPS \cite{abbas2021combinatorial}          & RN50      & 62.1  & -     & -    \\

Panoptic FCN \cite{li2021fully}           & RN50-FPN  & 61.4  & -     & -    \\
AdaptIS  \cite{sofiiuk2019adaptis}        & RN50      & 59.0  & 32.3   & 75.3  \\

\hdashline
Panoptic-DL \cite{cheng2020panoptic}       &  RN50     & 59.9  & 32.1    & 78.5    \\
\cellgray Panoptic-DL \cite{cheng2020panoptic} \textcolor{RoyalBlue}{+ OCR}   & \cellgray RN50     & \cellgray 60.7  & \cellgray 32.1   \cellgray & \cellgray 79.6    \\
\cellgray Panoptic-DL \cite{cheng2020panoptic} \textcolor{red}{+ PISR}     &  \cellgray RN50     & \cellgray {\textbf{62.2}}    & \cellgray {\textbf{33.3}}    & \cellgray {\textbf{80.2}} \\
\Xhline{1\arrayrulewidth}
AdaptIS  \cite{sofiiuk2019adaptis}         & RNX101  & 62.0   & 36.3    & 79.2  \\

UPSNet   \cite{xiong2019upsnet}      & RN101   & 61.8 & \textbf{39.0} & 79.2 \\

Panoptc FPN  \cite{kirillov2019fpn}  & RN101-FPN  & 61.2 & 36.7 & 80.4 \\

EfficientPS  \cite{mohan2020efficientps}   &    EffB5  & 63.9 & 38.3 & 79.3  \\

Panoptic-DL \cite{cheng2020panoptic}      &  RN101     & 60.5    & 33.7    & 79.0    \\

Panoptic-DL \cite{cheng2020panoptic}       &  X71     & 63.0    & 35.3    & 80.5    \\

\hdashline
Panoptic-DL \cite{cheng2020panoptic}      &  HR48     & 63.4    & 36.2    & 80.6    \\
\cellgray Panoptic-DL \cite{cheng2020panoptic}  \textcolor{red}{+ PISR}                                   &  \cellgray HR48     & \cellgray {\textbf{64.1}} &  \cellgray \textbf{37.6}    & \cellgray {\textbf{80.7}} \\

\Xhline{2\arrayrulewidth}
\end{tabular}
\vspace{-5pt}
\caption{Quantitative evaluation on the Cityscapes val set in terms of PQ, AP, and mIoU. We compare with existing methods using RN-50 (upper part) or other variants (lower part) as backbone. We also apply PISR to Panoptic-DeepLab with HRNet-w48 backbone and compare it with other existing methods which use larger backbones. We report the model performance for which the training is initialized with ImageNet pre-trained weights. 
RN, RNX, Eff, and HR48 refer to ResNet, ResNeXt, EfficientNet, and HRNet-w48, respectively. Gray rows are new models (ours) introduced in this paper. The best numbers in each section are highlighted in bold. }
\label{tab:city}
\vspace{-5pt}
\end{table}

\setlength{\tabcolsep}{5.5pt}
\begin{table}[t]
\scriptsize
\hspace{-3pt}
\begin{tabularx}{0.48\textwidth}{c|lcccc}
\Xhline{2\arrayrulewidth} 
 \textbf{Split} &\textbf{Method}  & \textbf{Backbone} & \textbf{PQ}   & \textbf{PQ}$^{th}$ & \textbf{PQ}$^{st}$ \\ 
\hline
\multirow{20}{*}{Val} & UPSNet \cite{xiong2019upsnet}          & RN50-FPN   & 42.5 & 48.5   & 33.4   \\ 
&AUNet \cite{li2019attention  }         & RN50-FPN   & 39.6 & \textbf{49.1}   & 25.2  \\ 
& CIAE (640) \cite{gao2021learning  }    & RN50-FPN   & 39.5 & 44.4   & 33.1   \\ 

& COPS \cite{abbas2021combinatorial}     & RN50      & 38.4  & 40.5     & 35.2 \\ 

& OANet \cite{liu2019end   }             & RN50-FPN   & 39.0 & 48.3   & 24.9   \\ 

& AdaptIS \cite{sofiiuk2019adaptis  }    & RN50     & 35.9 & 40.3   & 29.3  \\ 
& SSAP \cite{gao2019ssap}                & RN50     & 36.5 & 40.1   & 32.0   \\ 
& LPSNet  \cite{Hong_2021_CVPR}    & RN50     & 39.1 & 43.9   & 30.1   \\ 

\cdashline{2-6}
& Panoptic-FPN \cite{kirillov2019fpn}     & RN50-FPN     & 39.2 & 46.6   & 27.9   \\ 
& \cellgray Panoptic-FPN \textcolor{red}{+ PISR}   & \cellgray RN50-FPN     & \cellgray \textbf{42.7} & \cellgray \textbf{48.7}   & \cellgray \textbf{33.6}   \\ 

\cdashline{2-6}
& Panoptic-DL \cite{cheng2020panoptic}  & RN50     & 35.5 &   37.8     &  32.0  \\ 
& \cellgray Panoptic-DL \cite{cheng2020panoptic} \textcolor{RoyalBlue}{+ OCR}  & \cellgray RN50 & \cellgray 37.2  &  \cellgray 38.9  &  \cellgray   35.7   \\ 
& \cellgray Panoptic-DL \cite{cheng2020panoptic} \textcolor{red}{+ PISR}                    & \cellgray RN50     & \cellgray \textbf{38.8} &  \cellgray \textbf{40.6}      & \cellgray \textbf{36.2}   \\ 
\cdashline{2-6}
& Panoptic-DL \cite{cheng2020panoptic} & HR48     & 37.8 & -   & -  \\ 
& \cellgray Panoptic-DL \cite{cheng2020panoptic} \textcolor{red}{+ PISR}             & \cellgray HR48     & \cellgray \textbf{40.7} & \cellgray \textbf{42.6}   & \cellgray \textbf{37.7} \\ 
\cdashline{2-6}
&Panoptic-FCN~\cite{li2021fully} & Swin-L     & 52.1 & 58.5 & 42.3  \\
& Pan-SegFormer~\cite{li2021panoptic} & PvTv2-B5     & \textbf{54.1} & \textbf{60.4} & \textbf{44.6}   \\
& Pan-SegFormer~\cite{li2021panoptic} & PvTv2-B2     & 52.6 & 58.2 & 43.3  \\
& Max-DeepLab~\cite{wang2021max} & Max-L     & 51.1 & 57.0 & 42.2  \\
& MaskFormer~\cite{cheng2021maskformer} & Swin-L     & 52.7 & 58.5 & 44.0  \\

\cdashline{2-6}
& UPerNet~\cite{xiao2018unified} & Swin-L     & 50.3 & 55.7 & 42.1  \\
& \cellgray UPerNet~\cite{xiao2018unified} \textcolor{red}{+ PISR} & \cellgray Swin-L     & \cellgray \textbf{52.9} & \cellgray \textbf{58.9} & \cellgray \textbf{43.8}  \\

\hline
\multirow{5}{*}{Test} &Panoptic-FCN~\cite{li2021fully} & Swin-L     & 52.7 & 59.4 & 42.5  \\

 &Refine~\cite{ren2021refine} & RNX101-FPN  & 51.5 & 59.6 & 39.2  \\
 &Max-DeepLab~\cite{wang2021max} & Max-L     & 51.3 & 57.2 & 43.4  \\

\cdashline{2-6}
 & UPerNet~\cite{xiao2018unified} & Swin-L     & 50.9 & 56.7 & 42.3  \\
 & \cellgray UPerNet~\cite{xiao2018unified} \textcolor{red}{+ PISR} & \cellgray Swin-L & \cellgray \textbf{53.2} & \cellgray \textbf{59.2} & \cellgray \textbf{44.2}  \\
\Xhline{2\arrayrulewidth}

\end{tabularx}
\vspace{-5pt}
\caption{Quantitative evaluation on the COCO validation and test sets, in terms of PQ, PQ$^{th}$, and PQ$^{st}$.
RN and HR48 indicate ResNet and HRNet-w48, respectively. Gray rows are our models introduced in this paper. Best numbers are highlighted in bold.}
\label{tab:coco}
\vspace{-8pt}
\end{table}

\textbf{COCO:} Our results on COCO are summarized in Table~\ref{tab:coco}. We apply PISR to different base architectures with various backbones and compare with recent SOTA methods. It can be seen that by adding PISR, we significantly enhance the panoptic segmentation performance of the base model. For instance, for Panoptic-DeepLab with ResNet-50 backbone, PISR increases the PQ score considerably from 35.5 to 38.8, and outperforming the option of using OCR, which has a PQ score of 37.2. 

We also compare with the recent state-of-the art works on both COCO val and test-dev splits. We trained a baseline UPerNet model modified for the panoptic segmentation task. We then compare with an UPerNet model trained after adding the PISR block. We obtain a PQ score of 52.7, outperforming the baseline by 2.4 PQ. As seen from the numbers, our approach obtains comparable performance with recent SOTA methods on both val and test splits, and ranks amongst the top methods on the public leaderboard. 

\textbf{ADE20K:} We further evaluate PISR on ADE20K. It can be seen in Table~\ref{tab:ade} that by applying PISR to MaskFormer, which has SOTA performance on ADE20K, we improved its accuracy further. For instance, when using the ResNet-101 backbone, PISR significantly improves the PQ score from 35.7 to 37.0. 

\begin{figure}[t]
\center

\includegraphics[width=0.97\linewidth]{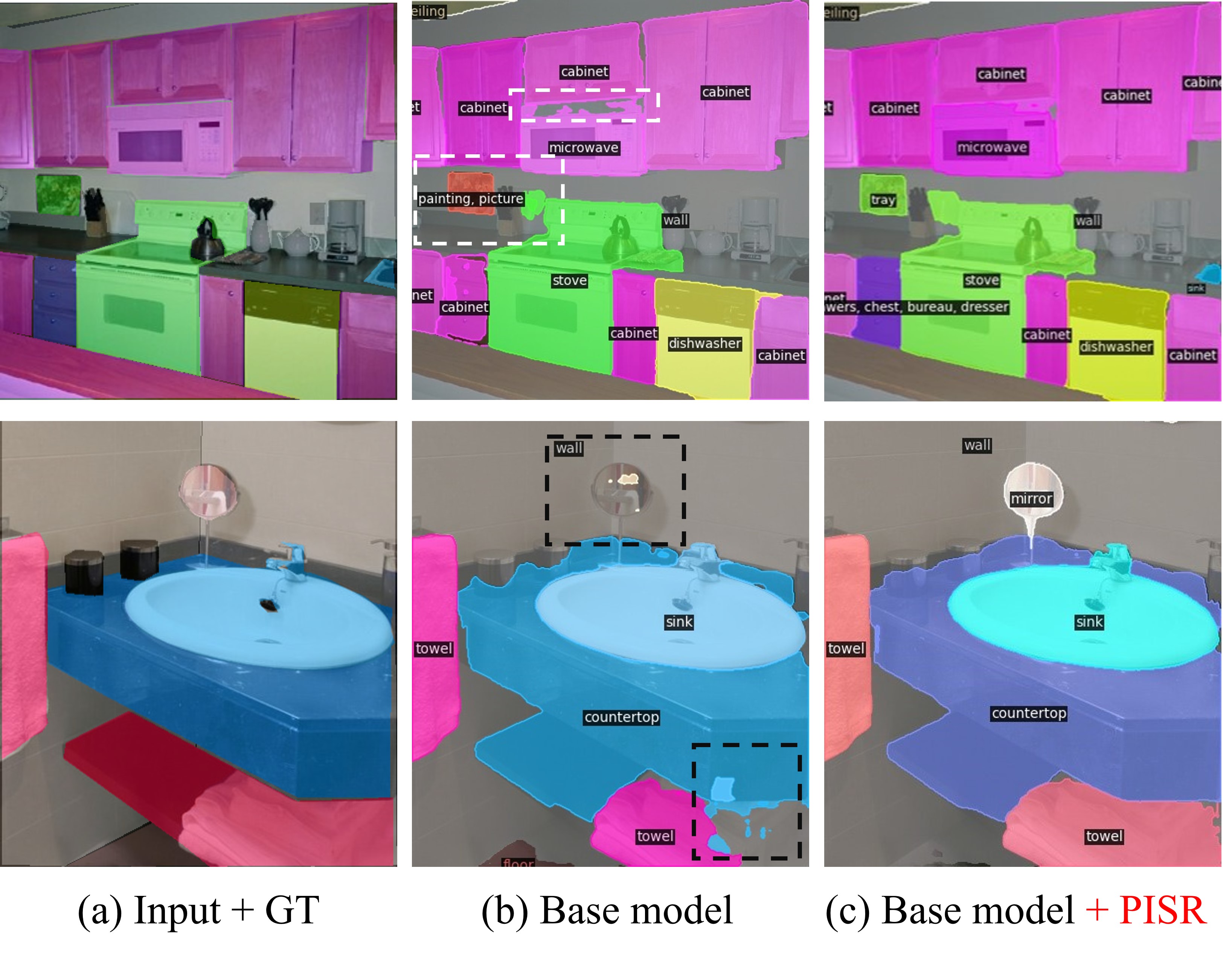}
\vspace{-8pt}
\caption{
Qualitative results \textbf{on ADE20K:}
(a) Input images with ground-truth masks overlaid.
(b) Predictions by using MaskFormer (ResNet-50).
(c) Our results by applying PISR to MaskFormer (ResNet-50).
When using PISR, the overall panoptic segmentation quality improves. Dashed boxes highlight sample regions where PISR significantly enhances the baseline prediction.}\label{fig:Visual results}
\vspace{-10pt}
\end{figure}

Figure~\ref{fig:Visual results} shows qualitative results obtained by using MaskFormer (ResNet-50) without and with PISR. Overall, PISR enhances the segmentation quality with clearer boundaries and more complete object masks (e.g., the towel in the bottom-row example). In addition, PISR has the capability to rectify masks that are entirely wrong. For instance, the tray in the top-row example is misclassified as a portrait when using the baseline alone (highlighted by a box), due to the image of people on it. By analyzing the instance relations, PISR successfully predicts that it is actually a tray. It is also visible that PISR can accurately segment the drawer, which was misclassified by the base model. 

\begin{table}[t]
\scriptsize
\centering
\begin{tabular}{lcccccc}
\Xhline{2\arrayrulewidth}
Method           & Backbone & PQ   & PQ$^{th}$ & PQ$^{st}$ & SQ & RQ \\ 

\hline
Panoptic-FCN \cite{li2021fully}    & RN50     & 30.1 & 34.1   & 27.3   & -    & -    \\ 

BGRNet \cite{wu2020bidirectional}    & RN50     & 31.8 & 34.1   & 27.3   & -    & -    \\ 
Auto-pan. \cite{wu2020auto}          & SV2    & 32.4 & 33.5   & 30.2   & -    & -     \\ 

\hdashline
MaskFormer \cite{cheng2021maskformer} & RN50     & 34.7 & 32.5   & 38.0   & 76.3 & 41.7 \\ 
\cellgray MaskFormer \textcolor{red}{+ PISR}                                & \cellgray RN50     & \cellgray 36.1 & \cellgray 34.7   & \cellgray 39.0   & \cellgray 78.3 & \cellgray 44.3 \\ 
MaskFormer & RN101    & 35.7 & 34.5   & 38.0   & 77.4 & 43.8 \\ 
\cellgray MaskFormer \textcolor{red}{+ PISR}                                & \cellgray RN101    & \cellgray \textbf{37.0} & \cellgray \textbf{35.6}   & \cellgray \textbf{39.7}   & \cellgray \textbf{79.9} & \cellgray \textbf{45.2} \\ 
\Xhline{2\arrayrulewidth}
\end{tabular}
\vspace{-5pt}
\caption{Quantitative evaluation on the ADE20K validation set. 
RN and SV2 indicates ResNet and ShuffleNetV2, respectively. Gray rows are new models (ours) introduced in this paper. Best numbers are highlighted in bold.}
\label{tab:ade}
\end{table}


\subsection{Accuracy-Computation Analysis}
\vspace{-2pt}

In Table~\ref{tab:gmacvspq}, we report the effect on computations (in GFLOPS) v/s accuracy (in PQ) by adding PISR for Maskformer and Panoptic-DeepLab respectively. We see that PISR increases baseline computations but provides a much better accuracy gain as compared to simply scaling up the backbone. For instance, when switching from ResNet-50 to ResNet-101, Maskformer increases its PQ from 34.7 to 35.7 on ADE20K, with GFLOPS increased from 116.6 to 159.3 \ds{and throughput decreased from 21.1 FPS to 19.6 FPS}. On the other hand, by adding PISR to Maskformer with ResNet-50, we achieve a higher PQ of 36.1, incur a smaller GFLOPS count of 136.0 \ds{and only slightly lower throughput of 20.4 FPS}. Similarly, for Panoptic-DeepLab, using PISR allows more accuracy gain and has less computation increase as compared to scaling up the backbone.

\begin{figure}[t]
\begin{center}
\includegraphics[width=1\linewidth]{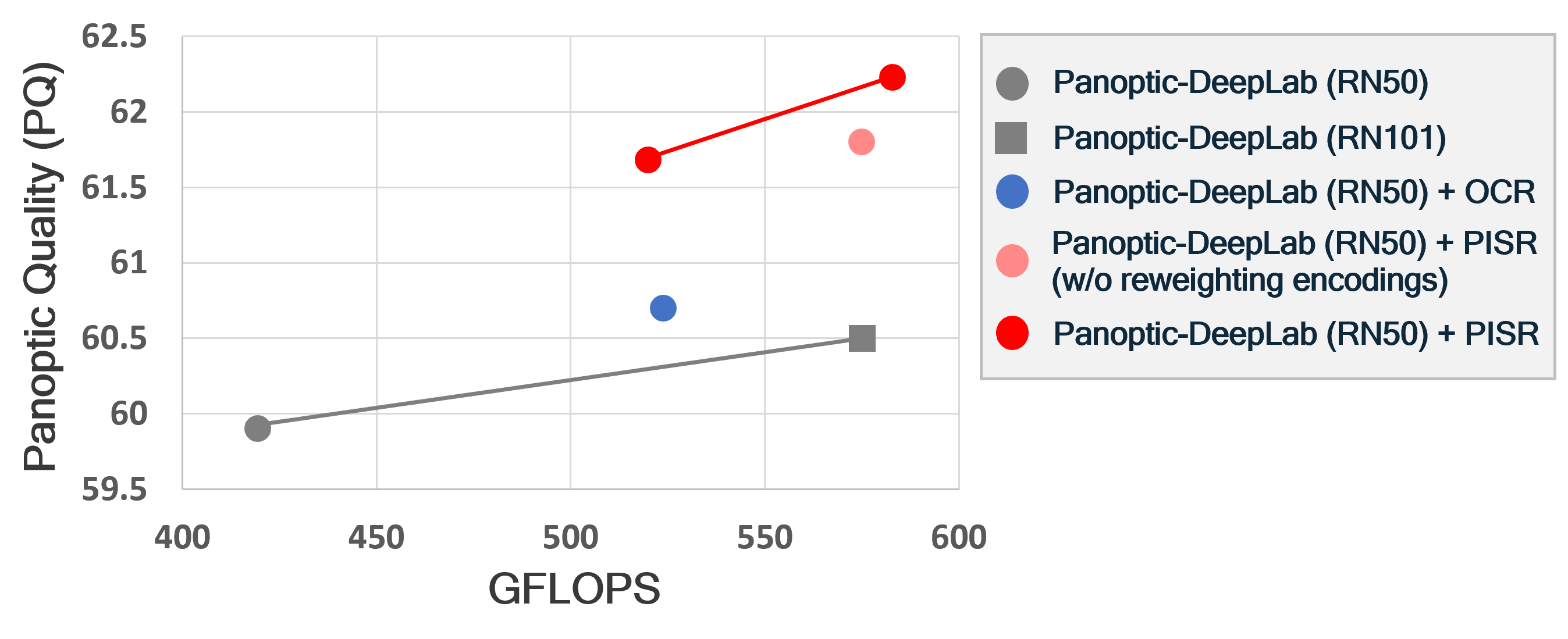}
   \end{center}
   \vspace{-20pt}
    \caption{Panoptic Quality vs. GFLOPS on Cityscapes, using Panoptic-DeepLab as the base model. We show results for the base model with different backbones, base model + OCR, base model + PISR (different model sizes), and base model + PISR without reweighting encodings. It can be seen that using PISR enables a much better accuracy-computation trade-off. 
    } 
    \label{fig: acc_mac}
    \vspace{-3pt}
\end{figure}

Figure~\ref{fig: acc_mac} provides a graphical illustration on using different ways to enhance panoptic segmentation accuracy by adding computation. The grey curve shows the effect of simply scaling up the baseline network. The blue dot shows the accuracy and computation resulted from applying the OCR module. The red curve shows the accuracy and computation of using PISR, by varying the intermediate number of channels in the PISR block. It can be seen that PISR provides a much better accuracy-computation trade-off as compared to scaling up the backbone or using OCR. We also show the effect of using PISR but without reweighting the encodings. While this has a lower accuracy and uses less computation, it still provides a favorable trade-off.

\strutlongstacks{T}
\begin{table}[t]
\scriptsize
\centering
\begin{tabular}{c|lcccc}
\Xhline{2\arrayrulewidth}
\textbf{Dataset}           & \textbf{Method} & \textbf{Backbone} & \textbf{PQ}   & \textbf{GFLOP} & \textbf{FPS} \\

\hline
\multirow{3}{*}{\Centerstack{ADE20K\\\tiny{($375X500$)}}} & MaskFormer \cite{cheng2021maskformer} & RN50     & 34.7 & 116.6 & \ds{21.1} \\ 
& \cellgray MaskFormer \textcolor{red}{ + PISR}                                & \cellgray RN50     & \cellgray 36.1 & \cellgray 136.0 & \cellgray \ds{20.4}\\ 
& MaskFormer & RN101    & 35.7 & 159.3 & \ds{19.6} \\ 
\hline
\multirow{3}{*}{\Centerstack{Cityscapes\\\tiny{($1024X2048$)}}} & Panoptic-DL \cite{cheng2020panoptic} & RN50     & 59.7 & 419.5 & 7.6 \\ 
& \cellgray Panoptic-DL \textcolor{red}{ + PISR}                                & \cellgray RN50  & \cellgray 62.2 & \cellgray 575.0 & \cellgray 6.8 \\ 
& Panoptic-DL & RN101 & 60.5 & 575.5 & 6.5 \\
\Xhline{2\arrayrulewidth}

\end{tabular}
\vspace{-5pt}
\caption{Accuracy (in PQ), computational complexity (in GFLOPS) \ds{and throughput (in FPS measured on Tesla-A100 GPU)} for Maskformer and Panoptic-DeepLab, on ADE20K val and Cityscapes val, respectively. It can be seen that using PISR provides more accuracy improvement and increases less computation as compared to simply using a heavier backbone.}
\label{tab:gmacvspq}
\vspace{-5pt}
\end{table}

\subsection{Ablation Studies}
\vspace{-3pt}

\begin{table}[t]
\scriptsize
\centering
\begin{tabular}{l cccc}
\Xhline{2\arrayrulewidth}
\textbf{Model}                           & \textbf{PQ}   & \textbf{mIoU} & \textbf{iIoU} & $\mathbf{\Delta}$   \\
\hline
Panoptic-DL \cite{cheng2020panoptic}                & 59.9 & 78.5 & 62.5 & 0       \\
\cellgray Panoptic-DL + Concatenation                         & \cellgray 60.6 & \cellgray 79.1 & \cellgray 62.6 & \cellgray + 1.4   \\
\cellgray Panoptic-DL + Elementwise Product                    & \cellgray 60.4 & \cellgray 78.2 & \cellgray 63.0 & \cellgray + 0.7   \\
\cellgray Panoptic-DL \textcolor{RoyalBlue}{+ OCR}  & \cellgray 60.7 & \cellgray 79.5 & \cellgray 62.4 & \cellgray + 1.7   \\
\cellgray Panoptic-DL \textcolor{magenta}{ + PISR (w/o Reweighting)}   & \cellgray 61.8 & \cellgray 79.6 & \cellgray 64.0 & \cellgray + 4.5   \\
\cellgray Panoptic-DL \textcolor{red}{ + PISR}   & \cellgray \textbf{62.2}    & \cellgray \textbf{80.2}    & \cellgray \textbf{64.4}    & \cellgray + \textbf{\ds{5.9}}        \\
\Xhline{2\arrayrulewidth}
\end{tabular}
\vspace{-5pt}
\caption{Comparing alternative ways to process semantic and instance features. \ds{$\Delta$ is the sum of all gains in PQ, mIoU, and iIoU}.
}
\label{tab:abl1}
\vspace{-10pt}
\end{table}

\textbf{Comparing with Other Feature Processing Schemes:}
We implement and compare with other ways of processing semantic and instance features to generate the final features for panoptic segmentation. In this study, we use Panoptic-DeepLab (ResNet-50) as the base model and report results on the Cityscapes dataset. We consider the following alternative options: 1) \textit{Concatenation:} We simply concatenate $S$, $I$, and $F$, and use the stacked features for the final prediction. 2) \textit{Elementwise Product:} $S$ and $I$ are first concatenated and passed through a convolutional layer to match the dimensions of $F$. The elementwise product of the resulting tensor and $F$ is used as the final features. 3) \textit{OCR~\cite{yuan2021segmentation}:} We apply the OCR module to the semantic features. The instance features cannot be used as OCR cannot handle a variable number of predicted instances. Table~\ref{tab:abl1} summarizes the results. It can be seen that concatenation, elementwise product, and OCR brings minor improvements over the baseline as they do not properly capture the semantic and instance relations. In contrast, PISR enables a larger performance gain by properly encoding the relations.

\begin{figure}[t]
\centering

\subfloat[Top-$K$ analysis on Cityscapes val]{
  \includegraphics[width=0.75\linewidth, keepaspectratio]{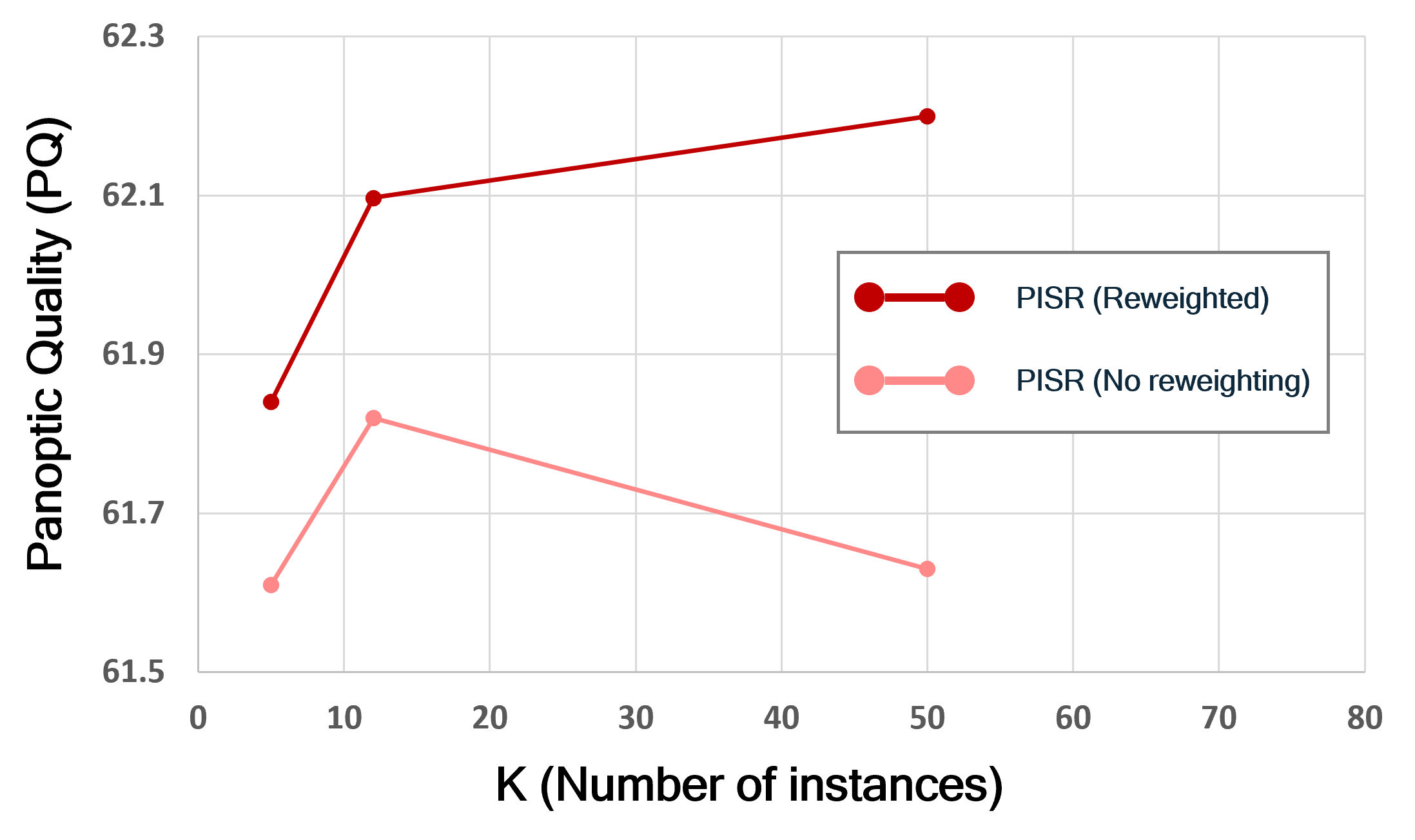}}
\vspace{0.2cm}
\subfloat[Top-$K$ analysis on COCO val]{
  \includegraphics[ width=0.75\linewidth, keepaspectratio]{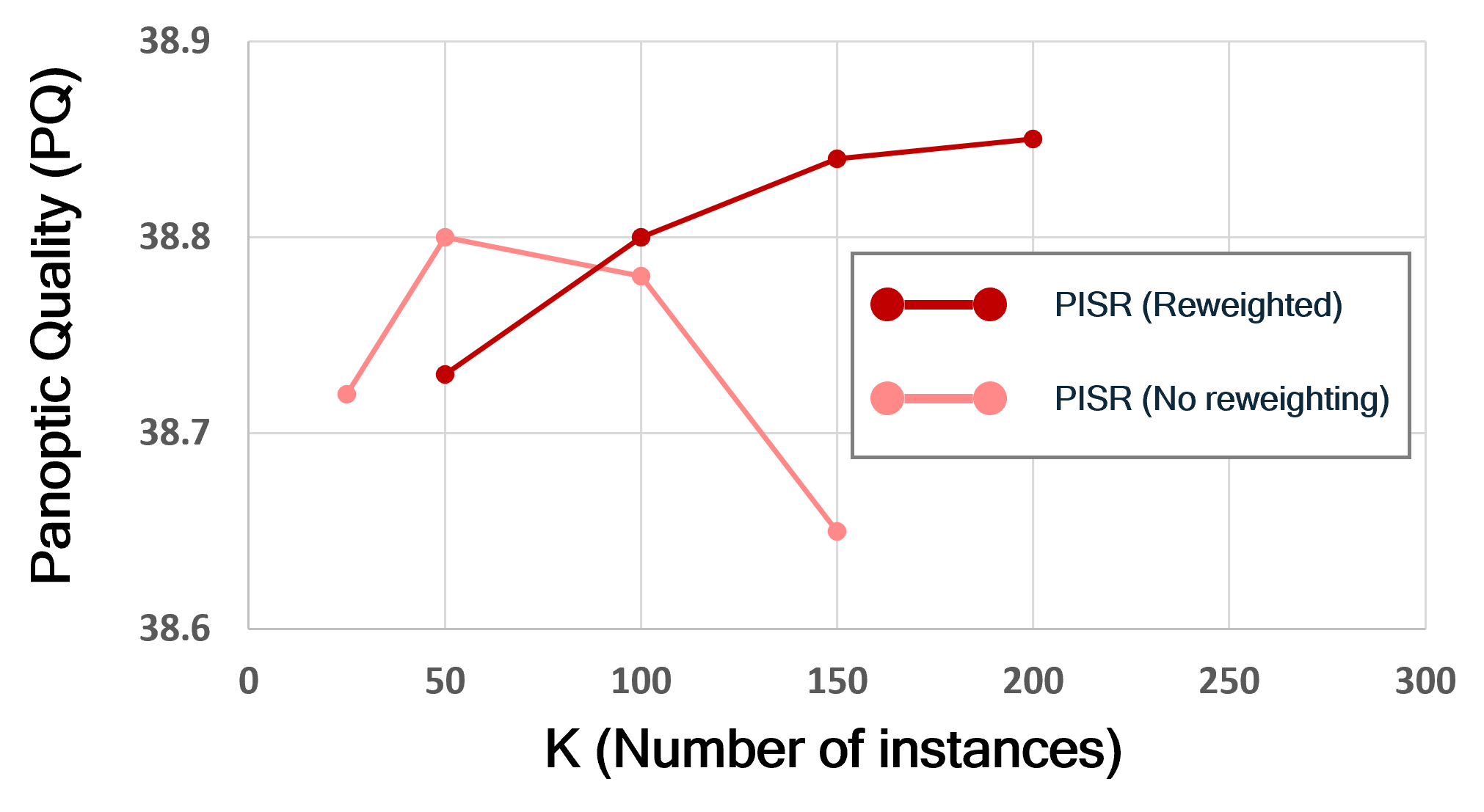}}
\vspace{-3pt}
    \caption{Observing PQ w.r.t. various values of $K$ on Cityscapes and COCO.}
    \label{fig: topk_coco}
    \vspace{-15pt}
\end{figure}

\textbf{Effect of $K$:} We study the effect of varying $K$ on the final PQ for both Cityscapes and COCO in Figure~\ref{fig: topk_coco}. The pink curves show the performance of PISR with Panoptic-DeepLab (ResNet-50). It can be seen that the PQ scores improve up to an optimum as $K$ increases, then the gain flattens out when $K$ becomes larger.

\textbf{Effect of Reweighting:} As shown in Table~\ref{tab:abl1}, applying our reweighting module provides additional gains compared to the case without reweighting. Although we use a small reweighting network, it provides a critical capability for PISR to focus on the important encodings rather than treating them all the same way. We apply the reweighting module at various $K$ and study the effect on PQ in Figure~\ref{fig: topk_coco}. As observed from the red curves, our reweighting module makes PISR more robust to $K$. PISR better learns to utilize only the useful information and does not suffer a heavy performance drop as $K$ increases (red curves in Figure~\ref{fig: topk_coco}). This also widens out the search-space greatly for $K$ as a hyperparameter. Without reweighting, the segmentation performance drops when $K$ becomes large since unreliable information might be permeated (pink curves in Figure~\ref{fig: topk_coco}).

\section{Conclusion}
\label{sec:conclusion}
\vspace{-5pt}
We propose a novel method called PISR to combine both semantic and instance information for panoptic segmentation. Our method consists of learning a general module that can be applied on top of existing panoptic segmentation architectures. The module captures interactions among instance and stuff embeddings to produce a context-aware segmentation map. Furthermore, a weighing scheme is introduced to produce varying contributions of different instances in the scene. Experimental evaluation on three large standard datasets for panoptic segmentation show that PISR improves segmentation performance over existing methods. We also conduct additional ablation studies and analyses to better understand our framework. 


{\small
\bibliographystyle{ieee}
\bibliography{egbib}

\begin{thebibliography}{10}\itemsep=-1pt

\bibitem{abbas2021combinatorial}
A.~Abbas and P.~Swoboda.
\newblock Combinatorial optimization for panoptic segmentation: A fully
  differentiable approach.
\newblock {\em arXiv preprint arXiv:2106.03188}, 2021.

\bibitem{bolya2019yolact}
D.~Bolya, C.~Zhou, F.~Xiao, and Y.~J. Lee.
\newblock Yolact: Real-time instance segmentation.
\newblock In {\em Proceedings of the IEEE/CVF International Conference on
  Computer Vision}, pages 9157--9166, 2019.

\bibitem{borse2021hs3}
S.~Borse, H.~Cai, Y.~Zhang, and F.~Porikli.
\newblock Hs3: Learning with proper task complexity in hierarchically
  supervised semantic segmentation.
\newblock 2021.

\bibitem{borse2021inverseform}
S.~Borse, Y.~Wang, Y.~Zhang, and F.~Porikli.
\newblock Inverseform: A loss function for structured boundary-aware
  segmentation.
\newblock In {\em Proceedings of the IEEE/CVF Conference on Computer Vision and
  Pattern Recognition}, pages 5901--5911, 2021.

\bibitem{Cai2021xdistill}
H.~Cai, J.~Matai, S.~Borse, Y.~Zhang, A.~Ansari, and F.~Porikli.
\newblock X-distill: Improving self-supervised monocular depth via cross-task
  distillation.
\newblock In {\em Proceedings of the British Machine Vision Conference}, 2022.

\bibitem{chen2014semantic}
L.-C. Chen, G.~Papandreou, I.~Kokkinos, K.~Murphy, and A.~L. Yuille.
\newblock Semantic image segmentation with deep convolutional nets and fully
  connected {CRFs}.
\newblock In {\em International Conference on Learning Representations}, 2015.

\bibitem{chen2017deeplab}
L.-C. Chen, G.~Papandreou, I.~Kokkinos, K.~Murphy, and A.~L. Yuille.
\newblock {DeepLab:} semantic image segmentation with deep convolutional nets,
  atrous convolution, and fully connected {CRFs}.
\newblock {\em IEEE transactions on pattern analysis and machine intelligence},
  40(4):834--848, 2017.

\bibitem{chen2019rethinking}
L.-C. Chen, G.~Papandreou, F.~Schroff, and H.~Adam.
\newblock Rethinking atrous convolution for semantic image segmentation. arxiv
  2017.
\newblock {\em arXiv preprint arXiv:1706.05587}, 2019.

\bibitem{chen2019tensormask}
X.~Chen, R.~Girshick, K.~He, and P.~Doll{\'a}r.
\newblock Tensormask: A foundation for dense object segmentation.
\newblock In {\em Proceedings of the IEEE/CVF International Conference on
  Computer Vision}, pages 2061--2069, 2019.

\bibitem{chen2020banet}
Y.~Chen, G.~Lin, S.~Li, O.~Bourahla, Y.~Wu, F.~Wang, J.~Feng, M.~Xu, and X.~Li.
\newblock {BANet:} bidirectional aggregation network with occlusion handling
  for panoptic segmentation.
\newblock In {\em Proceedings of the IEEE/CVF Conference on Computer Vision and
  Pattern Recognition}, pages 3793--3802, 2020.

\bibitem{cheng2020panoptic}
B.~Cheng, M.~D. Collins, Y.~Zhu, T.~Liu, T.~S. Huang, H.~Adam, and L.-C. Chen.
\newblock Panoptic-deeplab: A simple, strong, and fast baseline for bottom-up
  panoptic segmentation.
\newblock In {\em Proceedings of the IEEE/CVF conference on computer vision and
  pattern recognition}, pages 12475--12485, 2020.

\bibitem{cheng2021maskformer}
B.~Cheng, A.~G. Schwing, and A.~Kirillov.
\newblock Per-pixel classification is not all you need for semantic
  segmentation.
\newblock In {\em Advances in Neural Information Processing Systems}, 2021.

\bibitem{cordts2016cityscapes}
M.~Cordts, M.~Omran, S.~Ramos, T.~Rehfeld, M.~Enzweiler, R.~Benenson,
  U.~Franke, S.~Roth, and B.~Schiele.
\newblock The cityscapes dataset for semantic urban scene understanding.
\newblock In {\em Proceedings of the IEEE conference on computer vision and
  pattern recognition}, pages 3213--3223, 2016.

\bibitem{gao2019ssap}
N.~Gao, Y.~Shan, Y.~Wang, X.~Zhao, Y.~Yu, M.~Yang, and K.~Huang.
\newblock Ssap: Single-shot instance segmentation with affinity pyramid.
\newblock In {\em Proceedings of the IEEE/CVF International Conference on
  Computer Vision}, pages 642--651, 2019.

\bibitem{gao2021learning}
N.~Gao, Y.~Shan, X.~Zhao, and K.~Huang.
\newblock Learning category-and instance-aware pixel embedding for fast
  panoptic segmentation.
\newblock {\em IEEE Transactions on Image Processing}, 30:6013--6023, 2021.

\bibitem{ghiasi2016laplacian}
G.~Ghiasi and C.~C. Fowlkes.
\newblock Laplacian pyramid reconstruction and refinement for semantic
  segmentation.
\newblock In {\em European conference on computer vision}, pages 519--534.
  Springer, 2016.

\bibitem{he2017mask}
K.~He, G.~Gkioxari, P.~Doll{\'a}r, and R.~Girshick.
\newblock Mask r-cnn.
\newblock In {\em Proceedings of the IEEE international conference on computer
  vision}, pages 2961--2969, 2017.

\bibitem{he2016deep}
K.~He, X.~Zhang, S.~Ren, and J.~Sun.
\newblock Deep residual learning for image recognition.
\newblock In {\em Proceedings of the IEEE conference on computer vision and
  pattern recognition}, pages 770--778, 2016.

\bibitem{Hong_2021_CVPR}
W.~Hong, Q.~Guo, W.~Zhang, J.~Chen, and W.~Chu.
\newblock Lpsnet: A lightweight solution for fast panoptic segmentation.
\newblock In {\em Proceedings of the IEEE/CVF Conference on Computer Vision and
  Pattern Recognition (CVPR)}, pages 16746--16754, June 2021.

\bibitem{huang2019ccnet}
Z.~Huang, X.~Wang, L.~Huang, C.~Huang, Y.~Wei, and W.~Liu.
\newblock Ccnet: Criss-cross attention for semantic segmentation.
\newblock In {\em Proceedings of the IEEE/CVF International Conference on
  Computer Vision}, pages 603--612, 2019.

\bibitem{kirillov2019fpn}
A.~Kirillov, R.~Girshick, K.~He, and P.~Doll{\'a}r.
\newblock Panoptic feature pyramid networks.
\newblock In {\em Proceedings of the IEEE/CVF Conference on Computer Vision and
  Pattern Recognition}, pages 6399--6408, 2019.

\bibitem{kirillov2019panoptic}
A.~Kirillov, K.~He, R.~Girshick, C.~Rother, and P.~Doll{\'a}r.
\newblock Panoptic segmentation.
\newblock In {\em Proceedings of the IEEE/CVF Conference on Computer Vision and
  Pattern Recognition}, pages 9404--9413, 2019.

\bibitem{lee2020centermask}
Y.~Lee and J.~Park.
\newblock Centermask: Real-time anchor-free instance segmentation.
\newblock In {\em Proceedings of the IEEE/CVF conference on computer vision and
  pattern recognition}, pages 13906--13915, 2020.

\bibitem{li2018weakly}
Q.~Li, A.~Arnab, and P.~H. Torr.
\newblock Weakly-and semi-supervised panoptic segmentation.
\newblock In {\em Proceedings of the European conference on computer vision
  (ECCV)}, pages 102--118, 2018.

\bibitem{li2020unifying}
Q.~Li, X.~Qi, and P.~H. Torr.
\newblock Unifying training and inference for panoptic segmentation.
\newblock In {\em Proceedings of the IEEE/CVF Conference on Computer Vision and
  Pattern Recognition}, pages 13320--13328, 2020.

\bibitem{li2019attention}
Y.~Li, X.~Chen, Z.~Zhu, L.~Xie, G.~Huang, D.~Du, and X.~Wang.
\newblock Attention-guided unified network for panoptic segmentation.
\newblock In {\em Proceedings of the IEEE/CVF Conference on Computer Vision and
  Pattern Recognition}, pages 7026--7035, 2019.

\bibitem{li2021fully}
Y.~Li, H.~Zhao, X.~Qi, L.~Wang, Z.~Li, J.~Sun, and J.~Jia.
\newblock Fully convolutional networks for panoptic segmentation.
\newblock In {\em Proceedings of the IEEE/CVF Conference on Computer Vision and
  Pattern Recognition}, pages 214--223, 2021.

\bibitem{li2021panoptic}
Z.~Li, W.~Wang, E.~Xie, Z.~Yu, A.~Anandkumar, J.~M. Alvarez, T.~Lu, and P.~Luo.
\newblock Panoptic segformer.
\newblock {\em arXiv preprint arXiv:2109.03814}, 2021.

\bibitem{lin2014microsoft}
T.-Y. Lin, M.~Maire, S.~Belongie, J.~Hays, P.~Perona, D.~Ramanan,
  P.~Doll{\'a}r, and C.~L. Zitnick.
\newblock Microsoft {COCO}: Common objects in context.
\newblock In {\em European conference on computer vision}, pages 740--755.
  Springer, 2014.

\bibitem{liu2019end}
H.~Liu, C.~Peng, C.~Yu, J.~Wang, X.~Liu, G.~Yu, and W.~Jiang.
\newblock An end-to-end network for panoptic segmentation.
\newblock In {\em Proceedings of the IEEE/CVF Conference on Computer Vision and
  Pattern Recognition}, pages 6172--6181, 2019.

\bibitem{liu2018path}
S.~Liu, L.~Qi, H.~Qin, J.~Shi, and J.~Jia.
\newblock Path aggregation network for instance segmentation.
\newblock In {\em Proceedings of the IEEE conference on computer vision and
  pattern recognition}, pages 8759--8768, 2018.

\bibitem{liu2021swin}
Z.~Liu, Y.~Lin, Y.~Cao, H.~Hu, Y.~Wei, Z.~Zhang, S.~Lin, and B.~Guo.
\newblock Swin transformer: Hierarchical vision transformer using shifted
  windows.
\newblock {\em arXiv preprint arXiv:2103.14030}, 2021.

\bibitem{long2015fully}
J.~Long, E.~Shelhamer, and T.~Darrell.
\newblock Fully convolutional networks for semantic segmentation.
\newblock In {\em Proceedings of the IEEE conference on computer vision and
  pattern recognition}, pages 3431--3440, 2015.

\bibitem{minaee2021image}
S.~Minaee, Y.~Y. Boykov, F.~Porikli, A.~J. Plaza, N.~Kehtarnavaz, and
  D.~Terzopoulos.
\newblock Image segmentation using deep learning: A survey.
\newblock {\em IEEE Transactions on Pattern Analysis and Machine Intelligence},
  2021.

\bibitem{mohan2020efficientps}
R.~Mohan and A.~Valada.
\newblock Efficientps: Efficient panoptic segmentation.
\newblock {\em International Journal of Computer Vision (IJCV)}, 2021.

\bibitem{Porzi_2019_CVPR}
L.~Porzi, S.~Rota~Bul\`o, A.~Colovic, and P.~Kontschieder.
\newblock Seamless scene segmentation.
\newblock In {\em The IEEE Conference on Computer Vision and Pattern
  Recognition (CVPR)}, June 2019.

\bibitem{qi2021pointins}
L.~Qi, Y.~Wang, Y.~Chen, Y.-C. Chen, X.~Zhang, J.~Sun, and J.~Jia.
\newblock Pointins: Point-based instance segmentation.
\newblock {\em IEEE Transactions on Pattern Analysis and Machine Intelligence},
  2021.

\bibitem{ren2021refine}
J.~Ren, C.~Yu, Z.~Cai, M.~Zhang, C.~Chen, H.~Zhao, S.~Yi, and H.~Li.
\newblock Refine: Prediction fusion network for panoptic segmentation.
\newblock In {\em Proceedings of the AAAI Conference on Artificial
  Intelligence}, volume~35, pages 2477--2485, 2021.

\bibitem{sofiiuk2019adaptis}
K.~Sofiiuk, O.~Barinova, and A.~Konushin.
\newblock Adaptis: Adaptive instance selection network.
\newblock In {\em Proceedings of the IEEE/CVF International Conference on
  Computer Vision}, pages 7355--7363, 2019.

\bibitem{tian2020conditional}
Z.~Tian, C.~Shen, and H.~Chen.
\newblock Conditional convolutions for instance segmentation.
\newblock In {\em Computer Vision--ECCV 2020: 16th European Conference,
  Glasgow, UK, August 23--28, 2020, Proceedings, Part I 16}, pages 282--298.
  Springer, 2020.

\bibitem{wang2021max}
H.~Wang, Y.~Zhu, H.~Adam, A.~Yuille, and L.-C. Chen.
\newblock Max-deeplab: End-to-end panoptic segmentation with mask transformers.
\newblock In {\em Proceedings of the IEEE/CVF Conference on Computer Vision and
  Pattern Recognition}, pages 5463--5474, 2021.

\bibitem{wang2020deep}
J.~Wang, K.~Sun, T.~Cheng, B.~Jiang, C.~Deng, Y.~Zhao, D.~Liu, Y.~Mu, M.~Tan,
  X.~Wang, et~al.
\newblock Deep high-resolution representation learning for visual recognition.
\newblock {\em IEEE transactions on pattern analysis and machine intelligence},
  2020.

\bibitem{wang2020solo}
X.~Wang, T.~Kong, C.~Shen, Y.~Jiang, and L.~Li.
\newblock Solo: Segmenting objects by locations.
\newblock In {\em European Conference on Computer Vision}, pages 649--665.
  Springer, 2020.

\bibitem{wang2020solov2}
X.~Wang, R.~Zhang, T.~Kong, L.~Li, and C.~Shen.
\newblock Solov2: Dynamic and fast instance segmentation.
\newblock {\em Proc. Advances in Neural Information Processing Systems
  (NeurIPS)}, 2020.

\bibitem{wu2020bidirectional}
Y.~Wu, G.~Zhang, Y.~Gao, X.~Deng, K.~Gong, X.~Liang, and L.~Lin.
\newblock Bidirectional graph reasoning network for panoptic segmentation.
\newblock In {\em Proceedings of the IEEE/CVF Conference on Computer Vision and
  Pattern Recognition}, pages 9080--9089, 2020.

\bibitem{wu2020auto}
Y.~Wu, G.~Zhang, H.~Xu, X.~Liang, and L.~Lin.
\newblock Auto-panoptic: Cooperative multi-component architecture search for
  panoptic segmentation.
\newblock {\em Advances in Neural Information Processing Systems}, 33, 2020.

\bibitem{xiao2018unified}
T.~Xiao, Y.~Liu, B.~Zhou, Y.~Jiang, and J.~Sun.
\newblock Unified perceptual parsing for scene understanding.
\newblock In {\em Proceedings of the European Conference on Computer Vision
  (ECCV)}, pages 418--434, 2018.

\bibitem{xiong2019upsnet}
Y.~Xiong, R.~Liao, H.~Zhao, R.~Hu, M.~Bai, E.~Yumer, and R.~Urtasun.
\newblock Upsnet: A unified panoptic segmentation network.
\newblock In {\em Proceedings of the IEEE/CVF Conference on Computer Vision and
  Pattern Recognition}, pages 8818--8826, 2019.

\bibitem{yang2019deeperlab}
T.-J. Yang, M.~D. Collins, Y.~Zhu, J.-J. Hwang, T.~Liu, X.~Zhang, V.~Sze,
  G.~Papandreou, and L.-C. Chen.
\newblock Deeperlab: Single-shot image parser.
\newblock {\em arXiv preprint arXiv:1902.05093}, 2019.

\bibitem{yang2020sognet}
Y.~Yang, H.~Li, X.~Li, Q.~Zhao, J.~Wu, and Z.~Lin.
\newblock Sognet: Scene overlap graph network for panoptic segmentation.
\newblock In {\em Proceedings of the AAAI Conference on Artificial
  Intelligence}, volume~34, pages 12637--12644, 2020.

\bibitem{yuan2021segmentation}
Y.~Yuan, X.~Chen, X.~Chen, and J.~Wang.
\newblock Segmentation transformer: Object-contextual representations for
  semantic segmentation.
\newblock In {\em European Conference on Computer Vision (ECCV)}, volume~1,
  2020.

\bibitem{zhang2022auxadapt}
Y.~Zhang, S.~Borse, H.~Cai, and F.~Porikli.
\newblock Auxadapt: Stable and efficient test-time adaptation for temporally
  consistent video semantic segmentation.
\newblock In {\em Proceedings of the IEEE/CVF Winter Conference on Applications
  of Computer Vision}, pages 2339--2348, 2022.

\bibitem{zhang2022perceptual}
Y.~Zhang, S.~Borse, H.~Cai, Y.~Wang, N.~Bi, X.~Jiang, and F.~Porikli.
\newblock Perceptual consistency in video segmentation.
\newblock In {\em Proceedings of the IEEE/CVF Winter Conference on Applications
  of Computer Vision}, pages 2564--2573, 2022.

\bibitem{zhao2017pyramid}
H.~Zhao, J.~Shi, X.~Qi, X.~Wang, and J.~Jia.
\newblock Pyramid scene parsing network.
\newblock In {\em Proceedings of the IEEE conference on computer vision and
  pattern recognition}, pages 2881--2890, 2017.

\bibitem{zhao2018psanet}
H.~Zhao, Y.~Zhang, S.~Liu, J.~Shi, C.~C. Loy, D.~Lin, and J.~Jia.
\newblock Psanet: Point-wise spatial attention network for scene parsing.
\newblock In {\em Proceedings of the European Conference on Computer Vision
  (ECCV)}, pages 267--283, 2018.

\bibitem{zhou2017scene}
B.~Zhou, H.~Zhao, X.~Puig, S.~Fidler, A.~Barriuso, and A.~Torralba.
\newblock Scene parsing through ade20k dataset.
\newblock In {\em Proceedings of the IEEE conference on computer vision and
  pattern recognition}, pages 633--641, 2017.

\end{thebibliography}
}

\end{document}


\appendix
\appendixpage
\counterwithin{figure}{section}
\counterwithin{table}{section}



\input{Supp_Text/Supp_Intro}
\input{Supp_Text/Supp_Exp}
\input{Supp_Text/Supp_Fig}
\input{Supp_Text/Supp_method}
\input{Supp_Text/Supp_heat}
